\documentclass[letterpaper, 10 pt, conference]{ieeeconf}
\IEEEoverridecommandlockouts
\makeatletter
\let\NAT@parse\undefined
\makeatother

\usepackage{graphicx}
\usepackage{authblk}
\usepackage{url}
\usepackage{amsfonts}
\usepackage{amsmath}
\usepackage{balance}
\usepackage{cite}
\usepackage{float}
\usepackage{makecell}
\usepackage{booktabs}
\usepackage[table]{xcolor}
\usepackage{cuted}
\usepackage{placeins}
\usepackage{indentfirst}
\usepackage{caption}
\usepackage{titlesec}
\titlespacing*{\subsubsection}{0pt}{0.5em}{0em}

\let\labelindent\relax
\usepackage{enumitem}
\usepackage{hyperref}
\usepackage{cleveref}
\hypersetup{
    colorlinks=true,
    linkcolor=black,
    citecolor=black,
    urlcolor=black
}
\usepackage{amssymb}
\usepackage{xspace}
\usepackage{stfloats}
\usepackage{caption} 
\usepackage{subcaption} 
\usepackage{microtype}

\newcommand{\method}{\textsc{UniMem}\xspace}

\definecolor{StanfordCardinal}{HTML}{9F1C1F}
\definecolor{StanfordLightRed}{HTML}{F4E6E7}
\definecolor{StanfordMidRed}{HTML}{D8A6A7}
\usepackage[breakable, most]{tcolorbox}  
\tcbset{
  stanfordstyle/.style={
    colback=StanfordLightRed,      
    colframe=StanfordCardinal,     
    colbacktitle=StanfordMidRed,   
    coltitle=black,                
    boxrule=1pt,                   
    arc=3pt,
    left=6pt,
    right=6pt,
    top=5pt,
    bottom=5pt,
    enhanced,
    fonttitle=\bfseries,
  },
}

\title{\LARGE \bf
    UniMem: Unifying Visual and Textual Memory in VLAs for Non-Markovian Tasks
}

\author{Lars Osterberg$^{1}$, Maggie Wang$^{1}$, and Mac Schwager$^{1}$
\thanks{$^{1}$Stanford University, Stanford, CA, USA}
}

\begin{document}
\bstctlcite{IEEEexample:BSTcontrol}
\maketitle
\thispagestyle{empty}
\pagestyle{empty}

\begin{abstract}

While Vision-Language-Action (VLA) models have leveraged internet-scale pretraining and task-specific finetuning to achieve strong performance on long-horizon tasks, they often struggle with non-Markovian tasks that require memory. Existing approaches to memory typically involve additional Vision-Language-Models (VLMs) for long-term memory management, introducing a memory bottleneck and a fractured training pipeline. Conditioning on multiple historical frames can provide the VLA with access to more descriptive features of past scenes, but can degrade performance if frames are chosen at arbitrary, fixed intervals. To address these limitations, we present \method, a framework that unifies high-level visual and textual memory and low-level control under one backbone. \method employs an event classifier to initiate textual and visual memory updates, a keyframe encoder for dense visual memory, and a keyframe caching technique to minimize overhead during policy rollouts. We evaluate \method across five simulation and four hardware tasks targeting sequential and spatial memory, demonstrating that our unified, single-model system outperforms fixed-interval image sampling baselines (93.4\% vs. 68.2\%) in simulation and hierarchical VLM-VLA baselines (80.0\% vs. 43.5\%) in hardware, while offering faster inference and a simple training pipeline for easy adoption.
\textbf{Project website:} \url{https://losterberg3.github.io/unimem-vla/}

\end{abstract}

\section{Introduction}
While vision-language-action (VLA) models~\cite{brohan2023rt1, octo_2023, brohan2023rt2, black2025pi0, intelligence2026pi07steerablegeneralistrobotic, nvidia2025gr00tn1openfoundation, kim2024openvla} have achieved remarkable progress in robot manipulation, their success is largely confined to short-duration, Markovian tasks due to an absence of architectural memory. For instance, state-of-the-art policies~\cite{pi05_2025} struggle on rudimentary tasks such as picking up an object and returning it to its starting position~\cite{chung2026rethinking, lian2026intentvlashorthorizonintentmodeling, shi2026memoryvla}, or waiting for a human gesture to cue a task. We characterize this failure mode as \textit{perceptual aliasing---when multiple temporally disparate moments in a training dataset share similar observations but demand different actions depending on preceding events}. The simplest example of non-Markovianity due to perceptual aliasing is an arm moving forward and backward once along the same path.  In this case, the same image and proprioception data conditions two opposite actions (forward and backward) depending on the task phase. We identify two types of memory required for a VLA to solve such non-Markovian tasks: (i) sequential or progress-aware memory to determine the current phase in the task, and (ii) spatial memory to retain pertinent information about the scene's unobservable state. Motivated by human cognition and recent breakthroughs in vision-language models showing that temporal history is crucial for context-aware reasoning~\cite{chen2025longvila}, we pose the central question: \emph{Can conditioning VLAs directly on multiple memory modalities unlock robust performance in non-Markovian tasks?}

\begin{figure}[t]
    \centering
    \includegraphics[width=\linewidth, trim=20pt 26pt 20pt 29pt, clip]{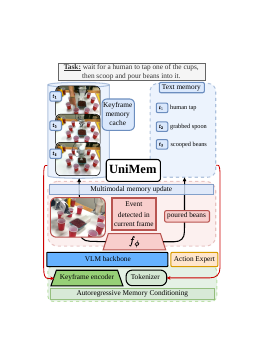}
    \caption{\small{\textbf{Overview of \method.} An event classifier ($f_\phi$) detects sparse sub-task transition events from the backbone latent space, autoregressively updating textual memory and a cache of encoded keyframe hidden states. This routes to our keyframe encoder and a tokenizer, providing event-driven memory directly to the backbone and action expert while unifying memory and control in one self-sustaining, low-latency VLA.
    }}
    \label{fig:overview}
    \vspace{-1.9em}
\end{figure} 

Existing methods~\cite{shi2026memoryvla, pmlr-v267-shi25d, sridhar2026memer, chen2026rmbenchmemorydependentroboticmanipulation} incorporate memory in VLA systems via an additional Vision-Language Model (VLM) that maintains textual summaries of completed milestones and produces sub-task commands for the VLA to execute. Although effective in simpler tasks, this factorized approach creates artificial silos that isolate the VLA from the rich historical context required for fine-grained reasoning. Furthermore, these approaches discard essential spatiotemporal information inherently encoded in past video frames---data that VLAs are already structured to process and act on.

To bridge this gap, we present \textbf{\method}, a unified memory framework that equips VLA models with direct access to historical context across both language and vision. \method is unified in two key respects: it integrates historical textual events with visual context, and it unifies memory and control under a shared backbone, as shown in~\Cref{fig:overview}. Crucially, \method positions the memory loop adjacent to the control loop to seamlessly integrate multiple forms of history, minimizing inference latency in non-Markovian tasks.

However, a longer historical context can also introduce spurious correlations \cite{pmlr-v305-villasevil25a, dehaan2019causal, spencer2021feedback}: a model that sees past observations regardless of their relevance will learn to exploit coincidental structure in the training data rather than the desired relationship between memory and action. In contrast to prior work, \method introduces an event-driven memory architecture that retains task-relevant textual and visual context to guide future action prediction. By separating meaningful history from task-irrelevant observations, this design mitigates memory-induced causal confusion and extends the VLA's temporal efficiency. Specifically, \method employs a lightweight event classifier that probes the VLA's latents to detect event transitions.  A detected event cues the system to store (i) an image keyframe, added autoregressively to a lightweight keyframe cache, and (ii) a textual description of the event added autoregressively to the text prompt. Since memory grows incrementally, we cache the transformer activations for keyframes---preserving real-time inference speeds by eliminating expensive attention-based computations over raw images. Furthermore, because frames are cached only when internal VLA features trigger the classifier head, our event detector serves a dual purpose as an in-distribution gating mechanism, ensuring the policy conditions exclusively on keyframes from high-confidence task milestones.

Our core contributions are as follows: 
\begin{itemize}
    \item \textbf{Textual and Visual Memory Conditioning:} We equip VLAs with unified memory across both textual and visual modalities for robust performance in challenging non-Markovian tasks. By co-locating the VLA's memory and control, \method can be trained in a single stage.
    \item \textbf{Event-Driven Architecture:} We introduce a lightweight event classifier that dynamically detects task-critical milestones conditioned on the VLA's latent space, and autoregressively constructs task-relevant memory from informative language and vision keyframes.
    \item \textbf{Real-Time Inference via Keyframe Caching:} By caching keyframes across inference steps, \method preserves single-frame inference speeds (${\sim} 90$\,ms) despite conditioning on rich visual histories, achieving a $6\times$ speedup over hierarchical memory baselines.
\end{itemize}

Through extensive simulation and hardware experiments (with 25 and 15 trials respectively), we demonstrate the robust performance of \method compared to state-of-the-art baselines, e.g., MemER~\cite{sridhar2026memer}. Across nine benchmark tasks, \method achieves approximately $93.4\%$ and $80.0\%$ average success rates in simulation and real-world hardware, respectively, compared to $72.6\%$ and $43.5\%$ success rates achieved by the best-performing baselines.

\section{Related Work}

\subsection{Recurrent \& Sequence-Based Robotic Memory}

To tackle partial observability and non-Markovian dynamics, early approaches incorporated temporal sequence models over continuous observation histories. Standard methods explored recurrent architectures like LSTMs and RNNs over temporal sequences~\cite{rahmatizadeh2018vision, pmlr-v164-mandlekar22a} or state-space belief encoders~\cite{liang2025membotmemorybasedrobotintermittent} to maintain implicit hidden states across time. Some methods retain memory through proprioception~\cite{zhang2025tavla}, although this is limited to tasks where past scene states are not necessary for future task decisions. Other frameworks use transformer-based architectures to ingest temporal sequences of frames~\cite{fang2019scene, pmlr-v235-lee24y, shafiullah2022behavior}, although the length of these sequences are limited and do not necessarily capture all task-pertinent moments due to fixed-interval sampling. While effective in low-dimensional or short-horizon domains, blindly ingesting dense observation streams scales poorly with horizon length—leading to severe computational overhead, context degradation, and an inability to selectively retain task-critical events.

\subsection{Decoupled \& Multimodal Memory in VLA Paradigms}

VLA models generalize across diverse robotic tasks by scaling up internet-scale vision-language pretraining to low-level motor control. While foundational works display remarkable semantic reasoning and zero-shot generalization to novel objects and scenes across different embodiments~\cite{black2025pi0, pi05_2025, nvidia2025gr00tn1openfoundation, wei2026psi0}, they provide actions based solely on the immediate camera frame and task instruction. Although highly effective in short-horizon tasks and fully observable decision processes, this reactive nature is fundamentally limiting in tasks that demand historical context. To address these limitations, recent works employ different memory representation and conditioning mechanisms to equip VLAs with long-horizon memory. 

For navigation, hybrid systems leverage a high-level VLM and a low-level VLA to construct topological environment graphs~\cite{pmlr-v270-xu25b}. For tabletop manipulation, frameworks like SAM2Act~\cite{fang2025sam2act} maintain spatial memory of past scene states via a latent memory bank. MemER~\cite{sridhar2026memer} saves historically relevant keyframes, but its decoupled architecture prevents the low-level actor from conditioning directly on memory. Alternatively, some works unify the hierarchical approach in one VLA for complicated multi-stage manipulation tasks~\cite{lin2026onetwovla}, with thinking and acting modes swapped intermittently; other recent methods condition the low-level policy on temporal image sequences~\cite{mark2025bpp, li2026cronusvla}. MEM~\cite{torne2026memmultiscaleembodiedmemory} introduces a memory architecture for VLAs that aligns more closely with prevailing paradigms. Specifically, MEM leverages a VLM to generate textual summaries of past events and subsequently assigns sub-tasks to the VLA. However, this approach perpetuates the artificial silos separating VLAs from critical historical context. Moreover, combining sub-task commands with fixed-interval historical frames degrades the efficiency of the visual memory context.

While these methods touch on many aspects of memory individually, none condition the VLA directly on memory from \textbf{multiple modalities}, under \textbf{one unified architecture}. In contrast to MEM and prior methods, \method unifies memory retrieval and action generation under a single backbone, enabling event-driven visual and textual conditioning while maintaining low-latency control. 

\section{Unified Memory for VLAs}

\subsection{Problem Formulation and Method Overview}
We consider a long-horizon, non-Markovian robotic manipulation task framed as a Partially Observable Markov Decision Process (POMDP). At each timestep $t$, the agent receives an observation of its environment $o_t$ and a high-level task instruction $g$. A standard reactive policy, which conditions exclusively on the current observation ($\pi(a_{t:t+H} | o_t, g)$), fails when tasks require historical context. We characterize this failure mode as \textit{perceptual aliasing}. Formally, perceptual aliasing occurs when two global steps $i$ and $j$ in a demonstration dataset yield near-identical observations (${o_i \approx o_j}$), but require different actions (${a_i \neq a_j}$) depending on task history. Resolving this ambiguity requires conditioning on a history of past states, mapping $o_{t-T:t}$ to the correct low-level motor action $a_t$.

\method addresses this problem by maintaining two complementary forms of history: textual memory ($\mathcal{M}_t$) that records discrete task events and visual keyframe memory ($\mathcal{H}_t$) that retains key spatial information. Both forms are updated online by an event classifier integrated directly into the VLA backbone and jointly condition subsequent action prediction.

\subsection{Event-Driven Multimodal Memory}
\label{sub:latent_event_classification}

We implement \method on top of $\pi_{0.5}$ \cite{pi05_2025}, an open-source VLA consisting of a PaliGemma backbone and a Gemma action expert \cite{beyer2024paligemma}. To compress execution history into discrete, task-relevant events, we define the event vocabulary ${\mathcal{E}_{\text{all}} = \mathcal{E} \cup \{ \text{null} \}}$, where $\mathcal{E}$ represents the subset of valid, task-critical milestones (e.g., \texttt{"grabbed box"}). As shown in Figure~\ref{fig:overview}, we append an \textit{event classifier head} $f_\phi$, implemented as an MLP, directly onto the final layer latent representation $z_t$ extracted from the VLA backbone. The classifier produces a probability distribution over event classes, and the predicted event $e_t \in \mathcal{E}_{\text{all}}$ at step $t$ is chosen via the most likely class index:
\begin{equation}
    e_t = \mathcal{E}_{\text{all}}\left[ \arg\max f_\phi(z_t) \right].
\end{equation}

We use $z_t$ for event prediction because it is the final-layer representation that would otherwise be used for autoregressive text generation in the underlying VLM. Rather than using this representation to predict a language token, we pass $z_t$ through the MLP event classifier $f_\phi$ to predict the event $e_t$. Because $f_\phi$ is trained jointly with the VLA, the event classification loss in Eq.~\eqref{eq:event_loss} acts as auxiliary supervision on the shared backbone used for action generation. This couples memory learning with control—rather than separating them across hierarchical modules—and prioritizes latent features that support both objectives.

At each step $t$, if the predicted class ${e_t \in \mathcal{E}}$ (i.e., ${e_t \neq \text{null}}$), the framework triggers a memory update and $e_t$ is dynamically appended in the form of natural language to our textual history $\mathcal{M}_t$, modifying the language instruction for the subsequent step: $\mathcal{M}_{t+1} = [\mathcal{M}_t \mathbin{\Vert} e_t]$. At the start of every rollout, we initialize $\mathcal{M}_t =\emptyset$. This tightly integrated loop provides explicit textual context that cleanly breaks the visual ambiguity of downstream states and anchors the model to its own task progress.

The same detection of a task-critical event also triggers a visual memory update. When ${e_t \in \mathcal{E}}$, \method stores the corresponding multi-view visual observation ${I = \{I^{\text{wrist}}, I^{\text{ext}}\}}$. These observations are appended to our collection of keyframes, defined as ${\mathcal{H}_t = \{ I_k \}_{k \in \mathcal{K}}}$, where $\mathcal{K}$ represents the set of discrete, non-consecutive event timestamps along with the current time step $t$. At the start of every rollout, we initialize $\mathcal{H}_t$ with only the initial scene image. Due to compute limitations during training, the size of this history is capped to three past milestones ($|\mathcal{K}| \le 4$), dropping the oldest frame when exceeding this limit during rollouts. Depending on event sparsity, this can equate to over a minute of visual memory. 

Crucially, the event classifier head $f_\phi$ operates on the unified latent representation from both textual memory $\mathcal{M}_t$ and visual history $\mathcal{H}_t$. Because $z_t$ is constructed via cross-attention over the tokens from both forms of memory, the event classifier inherently conditions on the entire multimodal history. This creates a recursive memory loop where prior milestones directly inform the detection of subsequent events. The updated policy formulation is thus defined as $\pi(a_{t:t+H}, e_t|\mathcal{H}_t, \mathcal{M}_t, q_t, g)$, where $q_t$ is proprioception.

\subsection{Efficient Keyframe Encoding and Caching}
\label{sub:keyframes-encoding}

To ingest this keyframe bank $\mathcal{H}_t$ without degrading execution speeds, we modify $\pi_{0.5}$'s vision encoder (SigLIP \cite{zhai2023sigmoid}). Specifically, we interleave causal temporal self-attention every few layers of the standard spatial attention stack; at these layers, the temporal attention pass reuses that same layer's own query/key/value and layer-norm weights, a configuration closely adapted from~\cite{torne2026memmultiscaleembodiedmemory, pmlr-v139-bertasius21a}. After the final layer, only the tokens for the current images, now containing causally relevant past information, condition the backbone to keep token-length fixed. Unlike prior approaches that sample frames at fixed temporal intervals and rely on separate pretraining, we train our keyframe encoder end-to-end during fine-tuning. Keyframes explicitly isolate semantically meaningful events rather than potentially uninformative fixed-interval frames, providing a concentrated signal for task-relevant memory representations without requiring a separate pretraining phase. 

To avoid repeatedly re-encoding historical keyframes, we use a cache of their intermediate representations, $\mathcal{H}_{\text{cache}}$, for future timesteps to attend to. Specifically, we cache their hidden states prior to the positional embeddings and layer-normalizations applied during each temporal attention step. When a new frame arrives, we simply shift the positional embedding applied to each cached entry without recomputing costly spatial attention at each layer; the current image alone forms the query for temporal attention. This allows the encoder to project dense spatiotemporal representations into the embedding of the VLA backbone while preserving a real-time control loop as temporal context scales.

\subsection{Data and Training}
\label{sub:data}

\begin{figure*}[t]
    \centering
    \includegraphics[width=\linewidth, trim=25pt 15pt 22pt 10pt, clip]{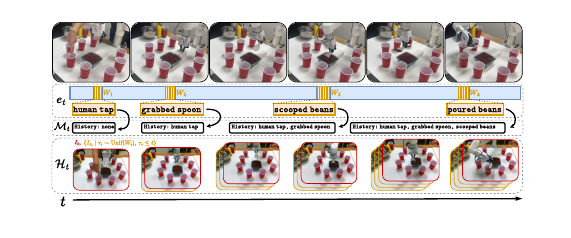}
    \caption{\small{\textbf{Overview of \method's automated data labeling procedure.} We prompt an agent to generate a script for the given task that labels demonstrations with $e_t$ and $\mathcal{M}_t$. Multiple frames in the window $W_i$ are labeled with the corresponding event, and $\mathcal{M}_t$ is only updated once this window has passed. At training time, a keyframe from each past $W_i$ along with $I_t$ is used to build $\mathcal{H}_t$.}}
    \label{fig:labeling}
    \vspace{-18pt}
\end{figure*}

To train our policies end-to-end, we need a robot action dataset (containing $a_{t:t+H}, o_t, q_t$, and $g$) labeled with $e_t$ and $\mathcal{M}_t$. Once demonstrations have been collected for a specific task, we prompt an agent (Claude Sonnet 5.0 \cite{Anthropic2026Claude}) to generate a script that parses our dataset for action signatures with the corresponding $e_t \in \mathcal{E}_{\text{all}}$ (i.e., a robot pitching upwards means scooping, the gripper closing means a bottle is being grasped, etc.). Frames outside these event windows get labeled with the null class by our script. For simplicity, we predefine our event vocabulary $\mathcal{E}_{\text{all}}$, although this could be constructed autonomously using methods like RoboInter \cite{li2026robointer} in future work. To prevent $\mathcal{M}_t$ from leaking the target during training, we only update $\mathcal{M}_t$ with an event once the entire window has passed. After the script is run on our dataset, a human verifies a subset of the demonstration videos has been labeled correctly and refines the script through more prompting if necessary. We build $\mathcal{H}_t$ at training time, sampling a single frame uniformly from each past event window along with $I_t$. Refer to Figure~\ref{fig:labeling} for a conceptual overview of this process.

With our dataset curated, we train the policy end-to-end with a two-term objective: the standard action-chunking loss and an auxiliary cross-entropy on the event classifier head.
We keep the flow-matching objective of $\pi_{0.5}$~\cite{pi05_2025} unchanged. The action expert integrates the chunk $a_{t:t+H}$ along the denoising velocity field conditioned on $o_t$; we denote this term $\mathcal{L}_{\text{a}}(\theta)$.
We supervise the event classifier head $f_\phi$ with a class-weighted cross-entropy loss against the automatically generated label $e_t$:
\begin{equation}
  \mathcal{L}_{\text{e}}(\theta, \phi) = -\,w(e_t)\,\log p_\phi(e_t \mid z_t)
  ~\label{eq:event_loss}
\end{equation}
where $w(e_t) = 1$ if $e_t \in \mathcal{E}$ and $w_{\text{null}}$ if $e_t = \text{null}$. Null frames dominate the dataset (most timesteps are not events), so we downweight them by setting $w_{\text{null}} = 0.02$, rather than masking them out of the loss entirely, since null frames provide a default for the head to predict if nothing eventful has happened. This prevents a wrong entry from being appended into $\mathcal{M}_t$ and $\mathcal{H}_t$ that would persist for the remainder of the rollout.
The two terms are optimized jointly,
\begin{equation}
  \mathcal{L} = \mathcal{L}_{\text{a}}
    + \lambda\,\mathcal{L}_{\text{e}},
  \qquad \lambda = 0.1.
\end{equation}
Since $z_t$ is a shared backbone representation rather than a detached
feature, the classification gradient flows back through the language-model
trunk, realizing the auxiliary supervision described in
Sec.~\ref{sub:latent_event_classification}.

\section{Experiments and Analyses}

\subsection{Experimental Setup}
\begin{figure}[t]
     \centering
     \includegraphics[width=\columnwidth, trim=26pt 15pt 23pt 15pt, clip]{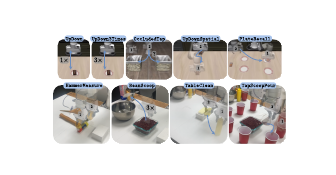}
     \caption{\small{\textbf{Overview of Tasks.} (Top) Manipulation tasks in \texttt{robosuite}. (Bottom) Real-world tasks on the xArm6 setup.}}
     \label{fig:tasks}
     \vspace{-18pt}
\end{figure}

We evaluate \method on nine memory tasks (shown in Figure~\ref{fig:tasks}), split across simulation and hardware. We use the \texttt{robosuite} environment with 7DoF Franka Panda for simulation~\cite{robosuite2020} and a UFactory xArm6 for hardware. In both domains, we equip our robot with wrist-mounted and external image streams. We use real time chunking for smooth inference on hardware \cite{NEURIPS2025_300ccb21}, querying \method at ${\sim} 10$~Hz. We retain up to three event keyframes ($|\mathcal{K}| = 3$) for simulation tasks and four ($|\mathcal{K}| = 4$) for hardware tasks. During experiments, a human observes each rollout, either from a saved video in simulation or in real time, and decides binary subtask success. 

\subsection{Baselines and Ablations}
We compare \method against VLA memory baselines and ablations on the memory input. In simulation, we compare against $\pi_{0.5}$ augmented with a video encoder, with frames sampled at 6-second intervals and processed using our keyframe encoder architecture ($|\mathcal{K}| = 4$). This baseline tests how arbitrary frame sampling performs relative to event-driven memory. 

On hardware, we compare against MemER with $\pi_{0.5}$ as the primary baseline, characterizing a hierarchical memory system~\cite{sridhar2026memer}. To isolate the contributions of textual and keyframe memory, we train separate ablations with no memory at all (vanilla $\pi_{0.5}$), text memory only ($I_t$ images only), and keyframe memory only (no $\mathcal{M}_t$). All ablations are trained with auxiliary classifier supervision.

\subsection{Simulation Experiments and Ablations}

We evaluate \method on five simulation tasks designed to assess its ability to address perceptual aliasing relative to fixed-interval visual history baselines across sequential and spatial memory settings:

\begin{itemize}[leftmargin=*]
    \item \texttt{UpDown}: Pick a box and put it down once.
    \item \texttt{UpDown3Times}: Pick a box and put it down three times.
    \item \texttt{OccludedTap}: Pick a box, place it in one of two bins, retract until the contents of both bins are occluded, and then tap the bin containing the box.
    \item \texttt{UpDownSpatial}: Pick up a box and place it back in its original location. Success is based on placement accuracy.
    \item \texttt{PlateRecall}: Pick a box from one of four plates, place it to the side, and tap the plate that originally had the box.
\end{itemize}

As shown in Table~\ref{tab:sim_results}, \method achieves a \textbf{93.4\%} simulation average across non-Markovian evaluation environments, substantially outperforming the fixed-window video baseline ($\pi_{0.5}\text{+V.E.}$, $68.2\%$). We first verify that our form of history conditioning overcomes the perceptual aliasing failure mode in \texttt{UpDown}. \method reaches $100\%$ success on this task, whereas $\pi_{0.5}$ struggles to distinguish whether it is picking up or putting down the box, only achieving $52\%$.

While visual-only memory conditioning struggles in counting tasks because of its fixed amount of tokens provided to the backbone, textual memory excels in these settings by compressing history into discrete events that grows incrementally. For instance, in \texttt{UpDown3Times}, keyframe-only conditioning fails ($23\%$) due to limited temporal reach, while text-only conditioning reaches $93\%$ success. 

In spatial memory tasks, joint visual and textual memory improves task performance. In \texttt{OccludedTap}, the keyframe ablation performs $12\%$ better than the text ablation; we hypothesize that this is because keyframes provide many memory pathways for the policy to exploit while text is left to just one. In continuous spatial grounding tasks (\texttt{UpDownSpatial}), text-only memory drops to $30\%$ without geometric awareness of past scenes. While the video encoder and keyframes offer better spatial guidance ($49\%$ and $52\%$ respectively), they still occasionally lose track of task progression; \method, which combines keyframes with textual memory, resolves both failure modes, improving the task success rate to $79\%$. Finally, in \texttt{PlateRecall}, text-only memory achieves $20\%$ success---roughly matching the 1-in-4 chance of correct plate selection. Keyframes and the video baseline match the full model ($96\%$), demonstrating that visual memory cleanly resolves discrete spatial ambiguities once progress is more easily inferred.

\begin{table}[t]
\centering
\footnotesize
\setlength{\tabcolsep}{6pt}
\renewcommand{\arraystretch}{1.1}
\caption{\textbf{Simulation Results \& Ablations.} Success rates (\%) ($N=25$). ($\dagger$) reports mean subtask success due to long-horizon complexity. ($\ddagger$) reports spatial accuracy.}
\label{tab:sim_results}
\begin{tabular}{l ccccc}
\toprule
\textbf{Task} 
  & \textbf{$\pi_{0.5} + \text{V.E.}$} 
  & \textbf{N.M.} 
  & \textbf{T.O.} 
  & \textbf{K.O.} 
  & \textbf{Ours} \\
\midrule
~~\textbf{\texttt{UpDown}}           & 84          & 52   & 96   & 92          & \textbf{100}  \\
~~\textbf{\texttt{UpDown3Times}}$^\dagger$ & 16          & 22   & 93   & 23          & \textbf{96}   \\
~~\textbf{\texttt{OccludedTap}}     & 96          & 60   & 88   & \textbf{100} & 96            \\
~~\textbf{\texttt{UpDownSpatial}}$^\ddagger$   & 49          & 6    & 30   & 52          & \textbf{79}   \\
~~\textbf{\texttt{PlateRecall}}     & \textbf{96} & 8    & 20   & \textbf{96} & \textbf{96}   \\
\midrule
\textbf{Avg. Performance}                & 68.2        & 29.6 & 65.4 & 72.6        & \textbf{93.4} \\
\bottomrule
\multicolumn{6}{l}{\tiny V.E. = Video Encoder, N.M. = No Memory, T.O. = Text Only, K.O. = Keyframe Only}
\end{tabular}
\vspace{-20pt}
\end{table}

\subsection{Real-world Experiments and Analyses}
\label{sub:results_analyses}

Our empirical evaluations of \method in real-world experiments explore the following three core research questions:
\begin{enumerate}[label=Q\arabic*)]
    \item Does providing textual and visual memory to the VLA overcome the memory bottleneck of hierarchical systems that only condition the VLA on subtask commands?
    \item How do textual and visual memories individually contribute to policy execution, and is joint conditioning necessary for task success?
    \item Does keyframe caching enable \method to maintain near single-frame inference latencies and scalability?
\end{enumerate}

We designed the following real-world tasks around these questions, testing both memory modalities and human interaction in long-horizon, non-Markovian settings:

\begin{itemize}[leftmargin=*]
    \item \texttt{HammerMeasure}: Measure the width of a hammer using a tape measure while controlling its retraction to prevent the hook from slipping from the hammer.
    \item \texttt{BeanScoop}: Pick up a spoon, put three scoops of beans into a bowl, and then place the spoon back.
    \item \texttt{TableClean}: Pick and place a bottle off a 60 cm by 80 cm table, pick up a sponge, wipe the bottle's original location, and then place the sponge back.
    \item \texttt{TapScoopPour}: Wait for a human to tap one of the eight cups, pick up a spoon, scoop the beans, pour the beans into the cup that the human tapped, and then place the spoon back.
\end{itemize} 

\begin{figure}[t]
    \centering
    \includegraphics[width=\columnwidth, trim=8pt 8pt 7pt 10pt, clip]{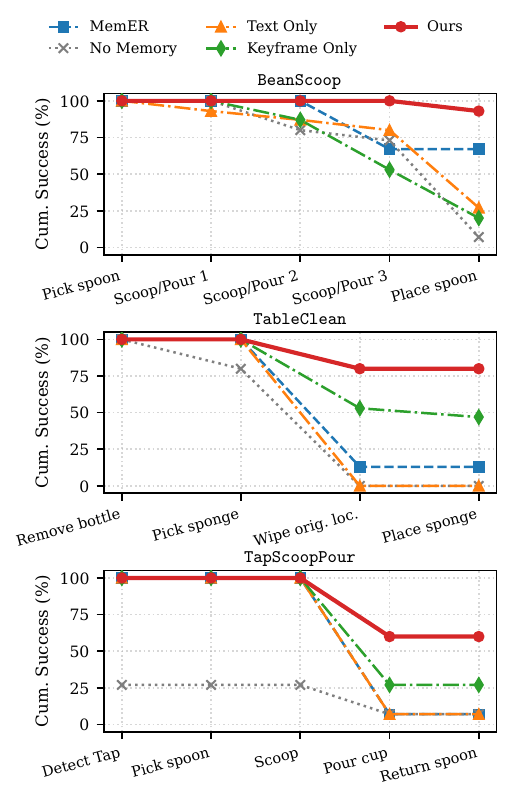}
    \caption{\small{\textbf{Cumulative Success Rates.} Hardware tasks of \texttt{BeanScoop}, \texttt{TableClean}, and \texttt{TapScoopPour}.}}
    \label{fig:task_breakdown}
    \vspace{-20pt}
\end{figure}

We also perform a simulated inference speed benchmark of \method to explore \textbf{Q3}. Overall success statistics on these tasks is reported in Table~\ref{tab:hardware_results}, while stage-by-stage analysis for select tasks is shown in Figure~\ref{fig:task_breakdown}. 

\begin{table}[t]
\centering
\footnotesize
\setlength{\tabcolsep}{6.7pt}
\renewcommand{\arraystretch}{1.1}
\caption{\small{\textbf{Hardware Results \& Ablations.} Success rates (\%) ($N=15$).}}
\label{tab:hardware_results}
\begin{tabular}{l ccccc}
\toprule
\textbf{Task} 
  & \textbf{MemER} 
  & \textbf{N.M.} 
  & \textbf{T.O.} 
  & \textbf{K.O.} 
  & \textbf{Ours} \\
\midrule
~~\textbf{\texttt{HammerMeasure}} & \textbf{87} & 13    & 53    & 53   & \textbf{87}   \\
~~\textbf{\texttt{BeanScoop}}     & 67          & 0    & 27    & 20    & \textbf{93}   \\
~~\textbf{\texttt{TableClean}}    & 13          & 0    & 0    & 47   & \textbf{80}   \\
~~\textbf{\texttt{TapScoopPour}}  & 7           & 7   & 7   & 27   & \textbf{60}   \\
\midrule
\textbf{Avg. Performance}        & 43.5        & 5.0   & 21.8   & 36.8   & \textbf{80.0} \\
\bottomrule
\multicolumn{6}{l}{\tiny N.M. = No Memory, T.O. = Text Only, K.O. = Keyframe Only}
\end{tabular}
\vspace{-6pt}
\end{table}

\subsubsection{Does \method overcome the memory bottleneck of hierarchical VLA systems?}

We test \method and the hierarchical baseline in tasks of graduated difficulty, ranging in the extent direct memory access is needed. In \texttt{HammerMeasure}, both \method and MemER achieve matching success rates (87\%), as high-level subtask commands are sufficient when task progression is simple and failure modes are purely mechanical due to the precision required to keep the hook on the hammer. However, as tasks grow in horizon and spatial complexity, the limits of hierarchical methods become evident.

In \texttt{BeanScoop}, MemER performs better than the ablations and proceeds to the third scoop in 67\% of rollouts. However, after completing two scoop-pour cycles, MemER's high-level planner may output an incorrect subtask command—instructing the robot to place the spoon when it should command a third scoop. By the time the high-level planner outputs a correction, the low-level policy is in an unrecoverable regime and ignores subsequent commands. In contrast, \method conditions the policy on memory directly at every control step. Continuous memory access prevents high-level misclassifications from happening in the first place, while near single-frame inference speeds ensure the robot continuously adjusts its trajectory before physical drift becomes uncorrectable, ultimately achieving \textbf{93.0\%} success.

In \texttt{TableClean}, we evaluate continuous spatial memory. Although MemER conditions the high-level policy on keyframes and augments its commands to the low-level policy with cues like left, right, and center, it only achieves 13\% success due to such a large area to select from. Only when keyframes and textual memory combine in full \method does the policy gain access to past visual states and wipe the bottle's exact spot on the table \textbf{80\%} of the time, while being off by no more than 10 cm during misses. 

In \texttt{TapScoopPour}, MemER pours into the correct cup 7\% of the time---even though MemER's command vocabulary is augmented, this is still insufficient to disambiguate such a large number of cups. Only when we provide visual history directly to the low-level policy do we see large improvements, with full \method achieving \textbf{60\%} success. Unified, multimodal memory provides spatial and sequential awareness directly to our VLA, an ability that hierarchical policies clearly lack.

\subsubsection{Is joint conditioning on textual and visual memory necessary for success?}

We compare our ablations in tasks of varying lengths and memory requirements to evaluate contributions of both memory modalities.
In \texttt{HammerMeasure}, we explicitly characterize perceptual aliasing. Without memory, the policy moves back and forth, struggling to distinguish task progress (13\%). Text-only and keyframe-only ablations improve to 53\%, but still struggle to distinguish and remember certain events. Only when both modalities combine in \method do we achieve robust memory (87\%).

In \texttt{BeanScoop}, with no memory, the policy doesn't stop scooping and pouring, collapsing on the most common signal from the dataset. With textual memory, long-horizon tracking improves to 27\%; however, the policy often moves to the next scoop prematurely without adding the previous pour to its memory, leading to undercounting. The keyframe-only ablation ameliorates this, only proceeding to the next scoop once a pour has been committed to memory. However, it cannot count all scoop-pour cycles with a limited context window and achieves 20\% success. When both modalities combine, we see robust long- and short-term memory, with \method achieving 93\% success.

In \texttt{TableClean}, with no memory, the policy fails to even begin wiping as the grabbing and placing actions alias each other. With only text, the policy wipes in random locations and does not perform a full wiping motion. With keyframes, the policy is endowed with spatial memory, completing the correct wipe 53\% of the time, but nonetheless suffers from infinite wipe sequences and fails to place the sponge. Only when keyframes and textual memory combine in full \method does the policy overcome these specific failure modes by gaining access to sequential and spatial memory.

In \texttt{TapScoopPour}, the policy without memory only proceeds to the grasp after the first tap 27\% of the time, requiring multiple taps for the rest. This is because as soon as the human stops tapping, the robot loses its signal to start. In the text ablation, the policy manages to pour into the correct cup 7\% of the time, essentially picking at random. Only with keyframes do we see an improvement, with that ablation achieving 27\% success. Its errors are also qualitatively different: misses are to an adjacent cup rather than arbitrary, indicating that the keyframe encoder does supply the correct spatial content. Adding textual memory further refines performance, with full \method achieving exact cup selection 60\% of the time. Across all tasks, neither the text- nor keyframe-only ablations matched the performance of full \method, demonstrating that textual and visual memory are complementary and essential for success on multimodal, non-Markovian tasks.

\subsubsection{Does \method maintain near single-frame inference speeds?}

\begin{figure*}[t]
    \centering
    \begin{subfigure}[b]{0.48\textwidth}
        \centering
        \includegraphics[width=\linewidth, trim=10pt 10pt 8pt 11pt, clip]{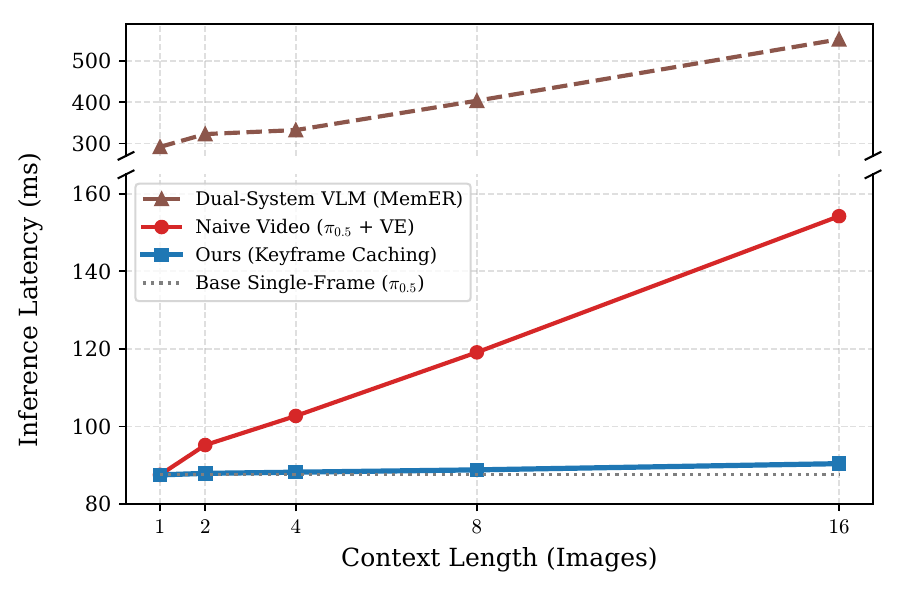}
        \label{fig:benchmark_a}
    \end{subfigure}
    \hfill
    \begin{subfigure}[b]{0.48\textwidth}
        \centering
        \includegraphics[width=\linewidth, trim=10pt 10pt 8pt 11pt, clip]{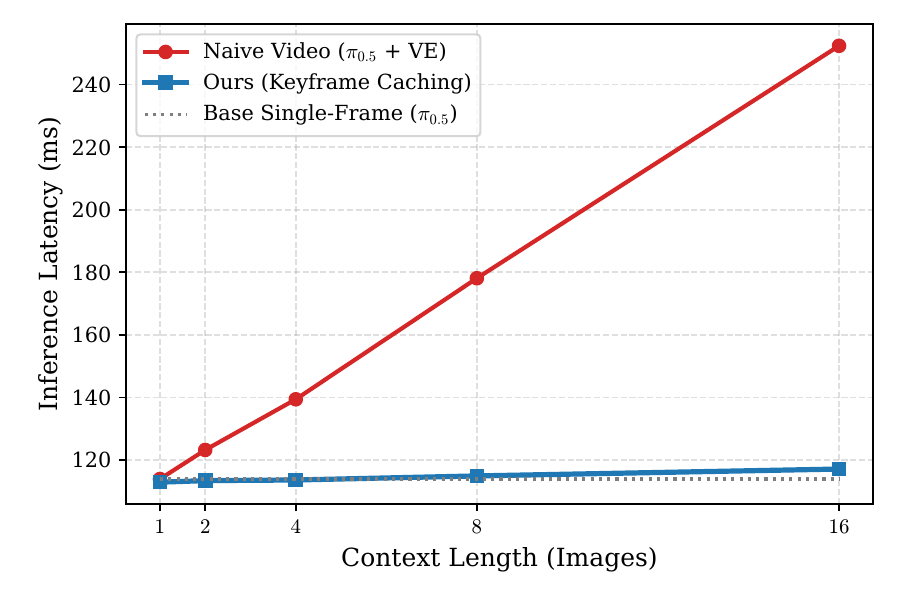}
        \label{fig:benchmark_b}
    \end{subfigure}
    \vspace{-1.8em}
    \caption{\small{Benchmarking inference latency on an RTX 4090. We compare inference speed when varying images per camera stream for 2 streams (left) and a simulated bimanual setup with 4 streams (right).}}
    \label{fig:benchmark}
    \vspace{-20pt}
\end{figure*}

To provide long-horizon memory with minimal computational overhead, \method leverages a hidden-state caching mechanism for keyframes. In a naive implementation, keyframes are retained as raw pixels and re-encoded through the vision backbone at every control step, leading to poor scaling as history grows. By instead caching the pre-computed keyframe representations prior to temporal self-attention, we eliminate redundant computations. 

We benchmark the latency of our caching mechanism against an ablation without caching, single-frame $\pi_{0.5}$, and MemER's dual-system VLM architecture across varying context window sizes for 2 and 4 camera streams to the left and right, respectively, of Figure~\ref{fig:benchmark}. Even on a standard workstation RTX 4090 GPU, maintaining a 16-keyframe memory context across four camera streams adds only ${\sim25}$ milliseconds of latency beyond the single-frame, 2-camera stream base policy. Although our experiments utilize contexts of up to 4 keyframes for 2-camera streams, these figures demonstrate that \method can scale to significantly longer context windows and different embodiments like a bimanual manipulator with 4-camera streams without sacrificing inference speed. Thus, in tasks significantly longer than the ones we evaluated \method on, the bottleneck would be training compute, which we leave for future work.

\section{Conclusion}

We present \method, a streamlined framework that unifies long-horizon spatial and sequential memory within a single VLA architecture. By using an event classifier to autoregressively update its history, \method avoids the memory bottleneck and high-latency of dual-system architectures. Across nine tasks in simulation and hardware, our system demonstrates superior task success, simpler training pipelines, and low-latency real-world rollouts.

Despite its strong empirical performance, our framework has several limitations that highlight promising directions for future research. First, while \method demonstrates robust state tracking over multi-minute execution horizons, it has not yet been evaluated on extended tasks spanning tens of minutes or hours. Exciting future work includes extending this framework with memory editing mechanisms—such as pruning and consolidation—to maintain high-performance. 

Second, our current pipeline relies on automated offline labeling. While this eliminates some human labor when curating the dataset, future work could leverage unsupervised temporal clustering or reinforcement learning to let the policy autonomously determine which events warrant retention. 

Finally, \method maintains memory with only one temporal context that extends several minutes. Distinguishing long-term from short-term memories to improve motor function and learn from mistakes, remains a promising direction for memory conditioning. Memory that lives through not only the input stream, but also in the weights of the policy itself, could make memory more natural and human-like.

\section*{ACKNOWLEDGMENT}

We used Claude Code to write some implementation code and our labeling scripts described in Section~\ref{sub:data}.

Toyota Research Institute provided funds to support this work. M. Wang is supported by the NASA NSTGRO Fellowship. 

\bibliographystyle{IEEEtran}
\bibliography{references}

\appendices

\section{Data Curation}
\label{app:hyperparameters}

We collect human teleoperated robot demonstrations at 10 Hz in simulation and 20 Hz on hardware. All labeling is done via a VLM-generated script, except for the human tap in \texttt{TapScoopPour} which is flagged whilst collecting the demonstration. An example prompt provided to Claude for the generation of labeling script for \texttt{TapScoopPour} is shown below. 

\begin{tcolorbox}[breakable, stanfordstyle, title=Example Claude Sonnet 5.0 Prompt]
\small
\setlength{\parskip}{0.5em}

Write a labeling function for our xArm demonstration dataset. The task being demonstrated is: \textbf{a human taps one cup, and the robot then puts a single scoop of beans into that cup.}

\textbf{Input.} One LeRobot parquet per episode, recorded at 20\,fps. Per frame you have the end-effector pose ($xyz$ in mm, roll/pitch/yaw), a gripper channel, and a \texttt{human\_event} column that is \texttt{1.0} on the single frame where the operator pressed the tap key during teleoperation. 

\textbf{Output.} The script should label each frame in the dataset with both a discrete event id from a vocabulary set and a textual memory string.

\textbf{Vocabulary.} Five events, each occurring exactly once per episode:
\texttt{\{0: "human tap", 1: "grabbed spoon", 2: "scooped beans", 3: "poured beans",
4: "placed spoon"\}}. Frames belonging to no event get the null target \texttt{-1}.

\textbf{Detection and labeling.}
\begin{itemize}\itemsep2pt
  \item \textbf{0 --- human tap:} first frame where \texttt{human\_event == 1.0}. Warn if
        there is more than one marker and use the first.
  \item \textbf{1 --- grabbed spoon:} the fully-closed gripper plateau of the first
        gripper close occurring after the tap.
  \item \textbf{2 --- scooped beans:} scan for the first
        40-frame window satisfying all of: roll std $< 5^\circ$, mean roll within
        $\pm 20^\circ$ of $180^\circ$ (either sign), yaw std $< 5^\circ$, mean
        $z < 255$\,mm, and pitch increasing on at least 60\% of frames (sustained
        straightening while low in the bowl). Label the end of the 40-frame window with this event.
  \item \textbf{3 --- poured beans:} the frame of minimum roll.
  \item \textbf{4 --- placed spoon:} first gripper open after the scoop.
\end{itemize}
Label the window of frames surrounding the event, starting 5 frames before and ending 20 frames after the detection. 

\textbf{Textual memory.} An event's phrase becomes visible in the memory string only
once the frame is no longer labeled with that event. Render as
\texttt{"History: human tap, grabbed spoon, \ldots"}, and \texttt{"History: none"} before
the first event is visible.
\end{tcolorbox}

\section{Training Details}
We fine-tune VLA checkpoints (MemER low-level $\pi_{0.5}$, $\pi_{0.5} + \text{V.E.}$, all ablations, and full \method) with LoRA \cite{hu2022lora}. For tasks such as \texttt{BeanScoop} and \texttt{TapScoopPour}, we note improved performance when upweighting the probability of sampling decisive moments by $\sim4\times$ in our training dataset. Conceptually, this focuses training effort on moments/frames where the robot's path diverges, such as deciding whether to scoop again or which cup to pour into. 

\end{document}